\documentclass[10pt,twocolumn,letterpaper]{article}

\usepackage{cvpr}              

\usepackage{graphicx}
\usepackage{amsmath}
\usepackage{amssymb}
\usepackage{booktabs}

\usepackage[pagebackref,breaklinks,colorlinks]{hyperref}

\usepackage[capitalize]{cleveref}
\crefname{section}{Sec.}{Secs.}
\Crefname{section}{Section}{Sections}
\Crefname{table}{Table}{Tables}
\crefname{table}{Tab.}{Tabs.}

\def\cvprPaperID{*****} 
\def\confName{CVPR}
\def\confYear{2022}

\begin{document}

\title{GT-Rain Single Image Deraining Challenge Report
}

\author{Howard Zhang\\
\and
Yunhao Ba\\
\and
Ethan Yang\\
\and
Rishi Upadhyay\\
\and
Alex Wong\\
\and
Achuta Kadambi\\
\and
Yun Guo\\
\and
Xueyao Xiao\\
\and
Xiaoxiong Wang\\
\and
Yi Li\\
\and
Yi Chang\\
\and
Luxin Yan\\
\and
Chaochao Zheng\\
\and
Luping Wang\\
\and
Bin Liu\\
\and
Sunder Ali Khowaja\\
\and
Jiseok Yoon\\
\and
Ik-Hyun Lee\\
\and
Zhao Zhang\\
\and
Yanyan Wei\\
\and
Jiahuan Ren\\
\and 
Suiyi Zhao\\
\and
Huan Zheng
}
\maketitle
\begin{abstract}
   This report reviews the results of the GT-Rain challenge on single image deraining at the UG2+ workshop at CVPR 2023. The aim of this competition is to study the rainy weather phenomenon in real world scenarios, provide a novel real world rainy image dataset, and to spark innovative ideas that will further the development of single image deraining methods on real images. Submissions were trained on the GT-Rain dataset and evaluated on an extension of the dataset consisting of 15 additional scenes. Scenes in GT-Rain are comprised of real rainy image and ground truth image captured moments after the rain had stopped. 275 participants were registered in the challenge and 55 competed in the final testing phase. 
\end{abstract}
\section{Introduction}
\label{sec:intro}

Natural weather conditions such as rain, fog, or snow can drastically reduce visibility, reduce contrast, and create noisy conditions that greatly degrade the quality of images of a scene. This naturally hurts the performance of many computer vision models and techniques designed and trained on clear image data~\cite{790306}. Running these models on this adverse data is a widely applicable problem, for example for self-driving cars, models that need to operate in geographical extremes, or models that must be highly robust~\cite{zhang2023perception}. Therefore, the field of single-image de-weathering, which focuses on techniques to remove weather effects from images, has grown significantly in popularity~\cite{10.1145/1360612.1360671, zhang2023weatherstream, ba2022not, wang2022uformer, Yang2017RainRemoval, valanarasu2022transweather, song2023vision, zhang2023data} After images are "de-weathered", general vision models can then be applied for various downstream tasks such as object detection, semantic segmentation, and more. 

Unfortunately, current single-image weathering techniques are bottlenecked by a common issue: dataset quality. It is not possible to obtain ideal real ground-truth pairs of rain and clean images because the same scene cannot possibly be observed both with and without rain at the same time. While previous works attempted to solve this problem by using simulated and pseudo-real pairs, these approaches lead to limited performance as current rain simulators cannot model all complex behaviours of rain or other weather effects.

For instance, a number of synthetic methods add rain streaks to clean images to generate dataset pairs, but rain does not only manifest as streaks: If raindrops are further away, the streaks meld together creating rain accumulation, or veiling effects, which are exceedingly difficult to simulate. A further challenge with generating high-quality synthetic data is that results on real test data can only be evaluated qualitatively, for no real paired ground truth exists. 

The GT-Rain dataset was proposed to tackle this problem by collecting high quality time multiplexed pairs of real rain and ground truth images. It consists of 31.5K time multiplexed pairs of real rainy and ground truth images captured moments apart in time. Each team must train their models using the GT-RAIN dataset but are free to use additional datasets to improve performance. Submissions are evaluated by comparing restored images against ground-truth images in the test dataset.


\section{GT-RAIN Challenge}
\label{sec:Single Image Derainig Challenge}

The GT-RAIN Challenge aims to study the rainy weather phenomenon in real world scenarios and to spark innovative ideas that will further the development of single image deraining methods on real images. It is built on top of the GT-Rain paper and dataset~\cite{ba2022not} and extends the GT-Rain dataset by collecting an additional 15 scenes used in the testing phase of the challenge to benchmark final submissions.

\subsection{The GT-RAIN dataset}
The GT-RAIN dataset was created by taking frames spaced closely in time and space, with and without rain. The key idea behind the data collection was to capture paired images on both sides of a key moment: the instant in which the weather effect stops (i.e. right before/after rain stops falling). Paired videos are collected within a short time frame of each other and then are strictly filtered to remove videos in which there is movement, illumination changes, camera motion, or other undesirable effects. Further correction algorithms are applied for subtle variations, such as slight movements of foliage. GT-RAIN features diverse and challenging scenarios that include (i) various types of rain conditions (i.e. long and short streaks, various densities and accumulation, with and without rain fog), (ii) large variety of background scenes from diverse urban locations (i.e. buildings, streets, cityscapes all across the world) to natural scenery (i.e. forests, plains, hills), (iii) varying degrees of illumination from different times of day, and all of which are captured by (iv) cameras that cover a wide array of resolutions, noise levels, and intrinsic parameters.

\begin{figure}
    \centering
    \includegraphics[width=0.4\textwidth]{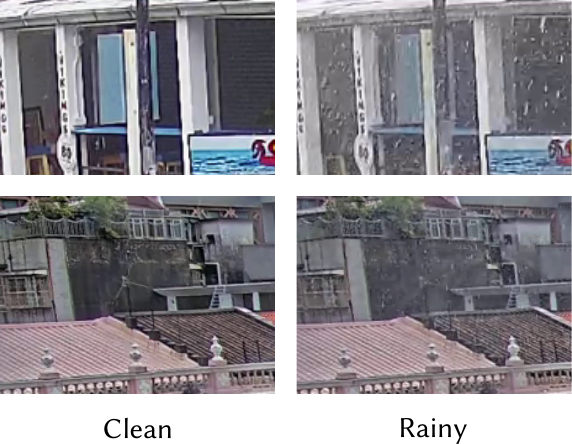}
    \caption{Example Images from the GT-RAIN test set.}
    \label{fig:example_data}
\end{figure}

\subsection{Evaluation}
\subsection{Challenge Phases}
The GT-RAIN challenge features three main stages
\begin{enumerate}
    \item \textbf{Training Phase} A training and validation set taken from GT-RAIN is provided to be used for development. Participants may also use any other dataset for training but must contain GT-RAIN in their final training set. The total training set includes a total of 76 scenes with 300 frames for each scene, for a total of 22,800 frames. Participants were also invited to use any method of there choosing, including data augmentations, ensembles, varying types of architectures, etc.
    \item\textbf{Validation Phase} Participants can optionally submit their results on the provided validation set to receive feedback on their latest standing on the leaderboard. This validation set includes a total of 14 scenes, for a total of 4200 frames.
    \item\textbf{Testing Phase} Participants must predict the derained result from a newly collected test set released exclusively in this phase, without access to the corresponding ground truth image. This test dataset was specifically collected for the purposes of the challenge and was not made public, including 15 scenes with 300 frames each, including images with varying rain effects (accumulation, streak sizes, streak shapes, etc.). Final scores are computed based on these results, using a combination of PSNR and SSIM metrics.
\end{enumerate}

\section{Challenge Results}

275 participants registered for the GT-Rain competition and 55 teams were evaluated in the final testing phase. Winning submissions were required to submit their code in addition to a report detailing their methods and submissions were evaluated using the standard PSNR and SSIM metrics. In particular, participants were ranked independently in terms of PSNR and SSIM and then those ranks were averaged to get the final rank of the submission. We now discuss some of the general trends and themes among submissions to the challenge and provide an in-depth analysis and results of the top 3 submissions.

\subsection{Architectures and General Trends}
In general, transformer based models were extremely popular, with most if not all teams using them. This is not particularly surprising, as transformers have grown significantly in popularity and most SOTA techniques across domains and fields are built on attention mechanisms. Therefore, in order to achieve the heavy compute requirements for transformers, most teams used GPUs for training and inference. Many of the teams chose to build their methods on top of existing de-raining models rather than starting from scratch. In particular, Restormer~\cite{Zamir2021Restormer}, its variants such as DRSformer~\cite{chen2023learning}, and Uformer~\cite{wang2022uformer} were popular choices to build off of.

In terms of training data, while many teams used GT-RAIN as their only training set, some teams opted to include additional synthetic datasts such as Rain200H~\cite{Yang2017RainRemoval}, Rain14K~\cite{fu2017removing}, and SPA-Data~\cite{wang2019spatial}. 

\section{Challenge Methods}
The top ranked teams in the GT-RAIN challenge both used entirely real datasets in their training phase. In addition to GT-RAIN, team HUST\_VIE added the WeatherStream dataset~\cite{zhang2023weatherstream}, also published by the GT-RAIN team, to their training phase. The other teams used only the GT-RAIN set. Both teams used transformer based models and built on top of previous models.

\subsection{Team HUST\_VIE}

\begin{figure*}
    \centering
    \includegraphics[width=0.9\textwidth]{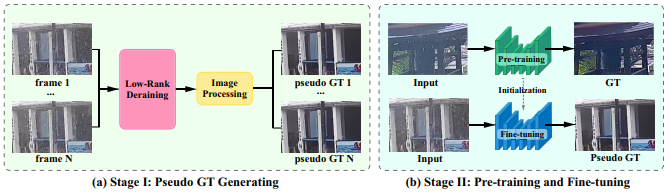}
    \caption{Overview of Team HUST\_VIE's technique for single image de-raining. Reproduced from~\cite{guo2023two}.}
    \label{fig:hust_res}
\end{figure*}

The top ranked team, Team HUST\_VIE, employed a two stage approach to the problem~\cite{guo2023two}. An overview of these stages is shown in~\cref{fig:hust_res}. In the first stage they utilize a low rank video deraining based technique~\cite{8100108} to generate a pseudo GT image for each input image. This pseudo GT is then passed through an image processing modules that moderates each pseudo GT output using contrast enhancement, by gamma correction, and sharpening, by subtracting the result of an unsharpening filter. 

In stage two of their process, team HUST\_VIE first pre-trains a Uformer model on a mixture of both GT-RAIN and WeatherStream rain datasets. They then fine-tune this model on the pseudo GT pairs generated in stage one. According to HUST\_VIE team, this approach allows them to provide a clear direction for pre-trained knowledge to transfer instead of simply depending on the generalizability of the model. They back this up with ablation studies which show an improvement of almost 2.5 points in PSNR and 0.03 points in SSIM for the fine-tuned model over just the pre-trained model.

Another interesting aspect of Team HUST\_VIE's work is that they designed their technique specifically based on the dataset being used. They first computed information such as pixel color statistics for varied types of scenes (e.g. cityscapes vs nature) and tested the alignment of provided frames. This gave them unique insights into the dataset that they could use to improve their techniques. For example, the pixel statistics told them that the GT-RAIN dataset contains significantly more rain-veiling effects than most de-raining datasets, allowing them to tailor their apporach to those problems.

Overall, Team HUST\_VIE ranked 2nd in PSNR and 1st in SSIM among all final submissions. As a result, they were the overall top team.

\subsection{Team FDL@ZLab}
\begin{figure*}
    \centering
    \includegraphics[width=0.9\textwidth]{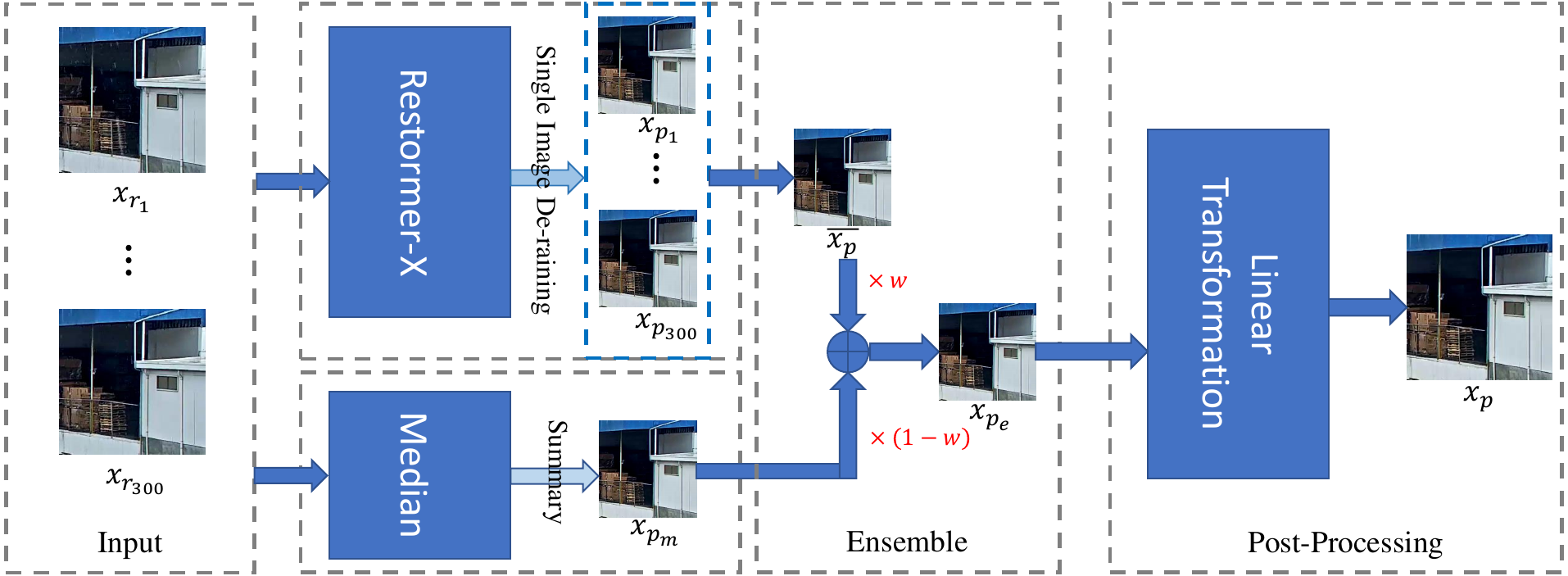}
    \caption{Overview of Team FDL@ZLab's technique for single image de-raining. Reproduced from~\cite{zheng2023restormer}.}
    \label{fig:fdl_overview}
\end{figure*}

Team FDL@ZLab implemented a Restormer-Plus approach consisting of four main modules~\cite{zheng2023restormer}. An overview is shown in Figure~\ref{fig:fdl_overview}

\begin{enumerate}
    \item \textbf{Initial de-raining module} This module, referred to as Restormer-X by the authors, consists of two submodules: the base Restormer model and a modified Restormer model. The modified model is very similar except that they replace the final convolutional layer with a fully connected layer designed for increased expressiveness. When running this module, they first determine whether a sample is "hard" or not by considering whether it contains features such as complex textures or drizzling rain. "Hard" samples are processed by the modified model while the easier ones are passed to the traditional Restormer.
    \item \textbf{The median filtering module} This module is built on the insight that rain artifacts are constantly changing, so a simple median filter applied to a sequence of rainy images can provide information about the underlying scene that might be missing from a single image. Therefore, this module performs median filtering on noisy scenes along the temporal direction to produce an estimated restored image.
    \item \textbf{The weighted averaging module} This module attempts to address potential overfitting in the Restormer-X module by taking a weighted average of the outputs from Restormer-X and the median filtering module. The weight for this average is a hyperparameter and was set at 0.9 weight for the Restormer-X image and 0.1 for the median filtered image.
    \item \textbf{The post-processing module} The final module adjusts the brightness of recovered images through a linear transformation: 
    $$x_{p,i} = \hat{a} \otimes x_{pe,i} + \hat{b}$$
    where $x_{p}$ is the final reconstructed image, $x_{pe}$ is the output of the weighted averaging module, and parameters $\hat{a}$ and $\hat{b}$ are estimated based on the following equations:
    \tiny
    $$\hat{a}_k = \frac{K \sum_{j \in I} (x_{pe,j,k} \times x_{est,j,k} - \sum_{j \in I} x_{pe,j,k} \sum_{j \in I} x_{est,j,k}}{K \sum_{j \in I} x_{pe,j,k}^2 - (\sum_{j \in I} x_{pe,j,k})^2}$$\\
    $$\hat{b}_k = \frac{\sum_{j \in I} x_{est,j,k}}{K} - \hat{a}_k \frac{\sum_{j \in I} x_{pe,j,k}}{K}$$
\end{enumerate}

Team FDL's process is able to achieve strong results through their use of multiple models for different data types, and various different processing methods whose results are all combined to output the final prediction.

Overall, Team FDL ranked 1st in PSNR and 4th in SSIM among all final submissions. They were therefore the 2nd overall team.

\section{Conclusion}

Rain is a weather effect that is both pervasive and damaging to countless computer vision applications. Unfortunately, it is a complex effect that can manifest itself in a multitude of shapes and sizes throughout a scene, dependent on camera parameters, scene illumination, or even geographic location. The purpose of the GT-RAIN Single Image Deraining Challenge is to inspire further efforts into creating algorithms that can remove the effect of rain on images. We describe in this challenge summary multiple different methods and trends that can be used for the de-raining task, and we hope the introduction of the GT-RAIN dataset will inspire many more in the future.

\appendix

\section{Participant List}

We list here the full author lists and affiliations of all teams who chose to provide their information.

\subsection{GT-RAIN Challenge Team}

\noindent
\textbf{\emph{Challenge:}}\\
GT-RAIN Challenge: Single Image Deraining for Real World Images

\noindent
\textbf{\emph{Organizers:}}\\
Howard Zhang\textsuperscript{1} (hwdz11508@ucla.edu), 
Yunhao Ba\textsuperscript{1} (yhba@ucla.edu),
Ethan Yang\textsuperscript{1} (eyang657@ucla.edu),
Alex Wong\textsuperscript{2} (alex.wong@yale.edu)

\noindent
\textbf{\emph{Affliations:}}\\
\textsuperscript{1} University of California, Los Angeles\\
\textsuperscript{2} Yale University\\

\subsection{Team HUST\_VIE}

\noindent
\textbf{\emph{Members:}}\\
Yun Guo\textsuperscript{1} (725452368@qq.com), Xueyao Xiao\textsuperscript{1}, Xiaoxiong Wang\textsuperscript{1}, Yi Li\textsuperscript{1}, Yi Chang\textsuperscript{1}, Luxin Yan\textsuperscript{1}

\noindent
\textbf{\emph{Affliations:}}\\
\textsuperscript{1} Huazhong University of Science and Technology\\

\subsection{Team FDL@ZLab}
\noindent
\textbf{\emph{Members:}}\\
Chaochao Zheng\textsuperscript{1} (zhengcc@zhejianglab.com), Luping Wang\textsuperscript{1}, Bin Liu\textsuperscript{1}

\noindent
\textbf{\emph{Affliations:}}\\
\textsuperscript{1} Research Center for Applied Mathematics and Machine Intelligence, Zhejiang Lab\\

\subsection{Team Avengers}
\noindent
\textbf{\emph{Members:}}\\
Sunder Ali Khowaja\textsuperscript{1} (sandar.ali@usindh.edu.pk), Jiseok Yoon\textsuperscript{2}, Ik-Hyun Lee\textsuperscript{2}\textsuperscript{3}

\noindent
\textbf{\emph{Affliations:}}\\
\textsuperscript{1} University of Sindh, Pakistan\\
\textsuperscript{2} IKLab Inc., South Korea \\
\textsuperscript{3} Tech University of Korea, South Korea

\subsection{Team LV\_Group\_HFUT}
\noindent
\textbf{\emph{Members:}}\\
 Zhao Zhang\textsuperscript{1} (cszzhang@gmail.com), Yanyan Wei\textsuperscript{1}, Jiahuan Ren\textsuperscript{1}, Suiyi Zhao\textsuperscript{1}, and Huan Zheng\textsuperscript{1}

\noindent
\textbf{\emph{Affliations:}}\\
\textsuperscript{1} Hefei University of Technology\\

{\small
\bibliographystyle{ieee_fullname}
\bibliography{egbib}
}

\end{document}